\documentclass[letterpaper, 10 pt, conference]{ieeeconf}  

\IEEEoverridecommandlockouts                              

\usepackage{amssymb,amsmath,amsfonts,mathrsfs}
\usepackage{graphicx}
\usepackage[table,xcdraw,dvipsnames]{xcolor}
\usepackage{tikz}
\usepackage{url}

\usepackage[utf8]{inputenc}

\usepackage[labelformat=simple]{subcaption}

\usepackage{colortbl}
\usepackage{booktabs}
\usepackage{array}
\usepackage{tabularx}

\usepackage[flushleft]{threeparttable}
\usepackage{multirow}

\let\labelindent\relax
\usepackage{enumitem}

\newcolumntype{L}[1]{>{\raggedright\let\newline\\\arraybackslash\hspace{0pt}}m{#1}}

\makeatletter
\DeclareRobustCommand\onedot{\futurelet\@let@token\@onedot}
\def\@onedot{\ifx\@let@token.\else.\null\fi\xspace}

\usepackage[T1]{fontenc}  
\usepackage[b]{esvect}    

\makeatletter
\newlength\xvec@height%
\newlength\xvec@depth%
\newlength\xvec@width%
\newcommand{\xvec}[2][]{%
  \ifmmode%
    \settoheight{\xvec@height}{$#2$}%
    \settodepth{\xvec@depth}{$#2$}%
    \settowidth{\xvec@width}{$#2$}%
  \else%
    \settoheight{\xvec@height}{#2}%
    \settodepth{\xvec@depth}{#2}%
    \settowidth{\xvec@width}{#2}%
  \fi%
  \def\xvec@arg{#1}%
  \def\xvec@dd{:}%
  \def\xvec@d{.}%
  \raisebox{.2ex}{\raisebox{\xvec@height}{\rlap{%
    \kern.05em
    \begin{tikzpicture}[scale=1]
    \pgfsetroundcap
    \draw (.05em,0)--(\xvec@width-.05em,0);
    \draw (\xvec@width-.05em,0)--(\xvec@width-.15em, .075em);
    \draw (\xvec@width-.05em,0)--(\xvec@width-.15em,-.075em);
    \ifx\xvec@arg\xvec@d%
      \fill(\xvec@width*.45,.5ex) circle (.5pt);%
    \else\ifx\xvec@arg\xvec@dd%
      \fill(\xvec@width*.30,.5ex) circle (.5pt);%
      \fill(\xvec@width*.65,.5ex) circle (.5pt);%
    \fi\fi%
    \end{tikzpicture}%
  }}}%
  #2%
}
\makeatother

\makeatletter
\renewcommand*\env@matrix[1][\arraystretch]{%
  \edef\arraystretch{#1}%
  \hskip -\arraycolsep
  \let\@ifnextchar\new@ifnextchar
  \array{*\c@MaxMatrixCols c}}
\makeatother

\definecolor{commentcolor}{gray}{0.5}
\usepackage{algorithm}
\usepackage{algpseudocode}
\algrenewcommand\algorithmicindent{1.0em}%

\algnewcommand{\LineComment}[1]{\State \textcolor{commentcolor}{\(\triangleright\) #1}}
\algnewcommand{\NewComment}[1]{\textcolor{commentcolor}{\(\triangleright\) #1}}
\algnewcommand{\To}{\textbf{to}}
\algnewcommand{\Break}{\textbf{break}}
\algnewcommand{\Continue}{\textbf{continue}}
\algnewcommand{\IIf}[1]{\State\algorithmicif\ #1\ \algorithmicthen}
\algnewcommand{\EndIIf}{\unskip}
\algnewcommand{\var}[1]{\textit{#1}}
\algnewcommand{\func}[1]{\textsc{#1}}

\def\publishedurl{\centering \textbf{{\color{red}{Workshop on The Role of Robotics Simulators for Unmanned Aerial Vehicles}}\\
{\color{red}{The 2023 International Conference on Robotics and Automation (ICRA), May 29 - June 2, 2023}}\\
\color{blue}{\url{https://imrclab.github.io/workshop-uav-sims-icra2023/}}}}

\title{\LARGE \bf
A Simulator for Fully-Actuated UAVs}

\author{Azarakhsh Keipour$^{1}$,
Mohammadreza Mousaei$^{2}$, and Sebastian Scherer$^{2}$
\thanks{$^{1}$ Amazon Robotics, Amazon, Washington, DC 20009, USA.
        {\tt\small keipour@gmail.com}}%
\thanks{$^{2}$ The Robotics Institute, Carnegie Mellon University, Pittsburgh, PA 15213, USA.
        {\tt\small \{mmousaei, basti\}@andrew.cmu.edu}}%
\thanks{* During the realization of this work, A. Keipour was affiliated with Carnegie Mellon University. The publication was written after A. Keipour joined Amazon.}%
    }

\begin{document}

\maketitle
\thispagestyle{empty}
\pagestyle{empty}

\begin{abstract}

This workshop paper presents the challenges we encountered when simulating fully-actuated Unmanned Aerial Vehicles (UAVs) for our research and the solutions we developed to overcome the challenges. We describe the ARCAD simulator that has helped us rapidly implement and test different controllers ranging from Hybrid Force-Position Controllers to advanced Model Predictive Path Integrals and has allowed us to analyze the design and behavior of different fully-actuated UAVs. We used the simulator to enable real-world deployments of our fully-actuated UAV fleet for different applications. The simulator is further extended to support the physical interaction of UAVs with their environment and allow more UAV designs, such as hybrid VTOLs. The code for the simulator can be accessed from
\url{https://github.com/keipour/aircraft-simulator-matlab}.

\end{abstract}


\section{Introduction} \label{sec:intro}

The rapid growth of Unmanned Aerial Vehicle (UAV) systems has resulted in reduced costs and increased researchers' attention to UAV development. New applications are introduced every day for free flight~\cite{Balaram2021, en15010217, Keipour:2022:sensors:mbzirc, RAKHA2018252}, and new designs are introduced (e.g., fully-actuated multirotors and hybrid VTOLs~\cite{Mousaei:2022:iros:vtol}) to allow even more applications that extend to aerial manipulation and physical interaction of these robots with their environment~\cite{Park2018a, Rashad2019a, Trujillo2019, Ollero2018, Keipour:2020:arxiv:integration}.

Many simulators have been introduced to support various aspects of the growth of UAV applications and research. The most common in use include Gazebo~\cite{2012simpar_meyer}, which is repurposed for UAVs, and AirSim~\cite{airsim2017fsr}, which focuses on the photo-realistic camera input. Add-ons such as RotorS~\cite{Furrer2016} have been developed that enhance these simulators. On the other hand, flight simulators, such as FlightGear~\cite{perry2004flightgear} and XPlane~\cite{x_2022}, have provided APIs for researchers which allow defining and programming aircraft, but their support for small aircraft (e.g., multirotors) is minimal. Defining new aircraft types (e.g., fully-actuated UAVs) on all these simulators is highly time-consuming, and the support for rapid prototyping and analysis of new designs is minimal. Furthermore, their support of physical interaction of UAVs is generally limited and requires many ``hacks'' to even visually appear natural. 

While developing a fully-actuated multirotor in our research, we realized that no existing simulator could help us rapidly analyze and test different aircraft designs or facilitate the development of new controllers for these new aircraft. As a result, we developed the ARCAD (AirLab Rapid Controller and Aircraft Design) simulator~\cite{Keipour:2023:scitech:simulator, Keipour:2022:thesis}, which allowed us to rapidly model and analyze capabilities of new aircraft, design and analyze new controllers, and implement and visualize new applications for them, including physical interaction applications. This simulator assisted us with aircraft ranging from fixed-pitched fully-actuated hexarotor to variable-pitch hybrid VTOL, with designing various controllers ranging from Hybrid Force-Position Controller (HPFC) to Model Predictive Path Integral (MPPI), and with UAV applications such as writing a text on a wall with a controlled force.

Figure~\ref{fig:features} illustrates some of the visualized features of the simulator. The major advantages of the simulator can be summarized below:

\begin{figure}[!t]
\centering
    \begin{subfigure}[b]{0.33\linewidth}
        \includegraphics[width=\textwidth]{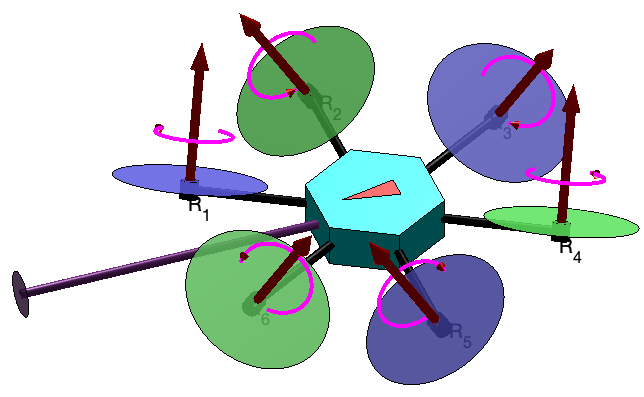}
        \caption{~}
    \end{subfigure}%
    \hfill%
    \begin{subfigure}[b]{0.66\linewidth}
        \includegraphics[width=\textwidth,trim={0 11cm 0 0},clip]{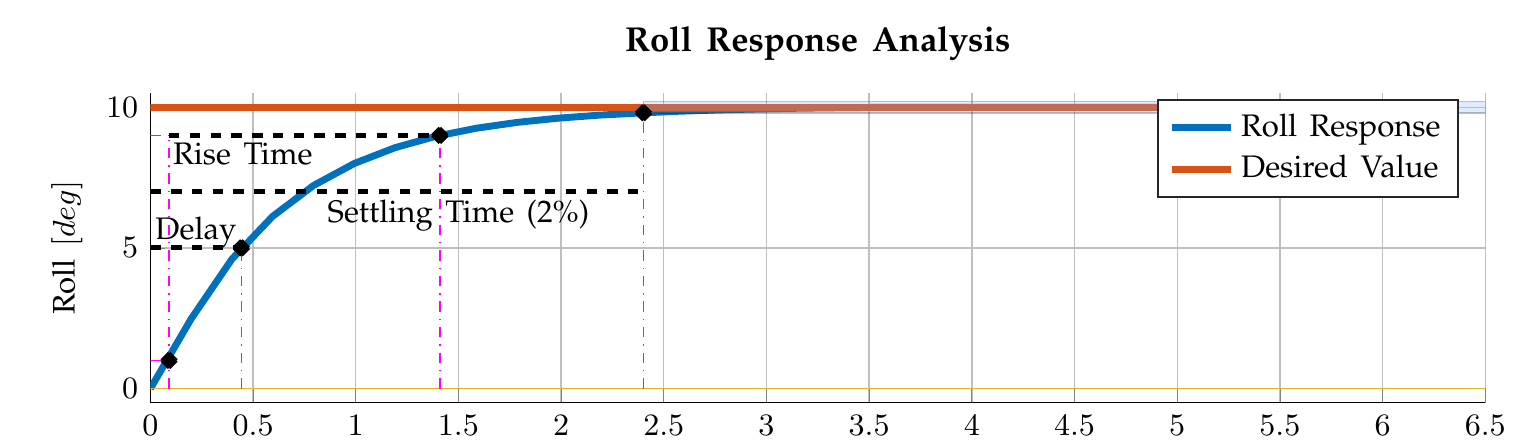}
        \caption{~}
    \end{subfigure}%
    
    \medskip
    \begin{subfigure}[b]{0.22\linewidth}
        \includegraphics[width=\textwidth]{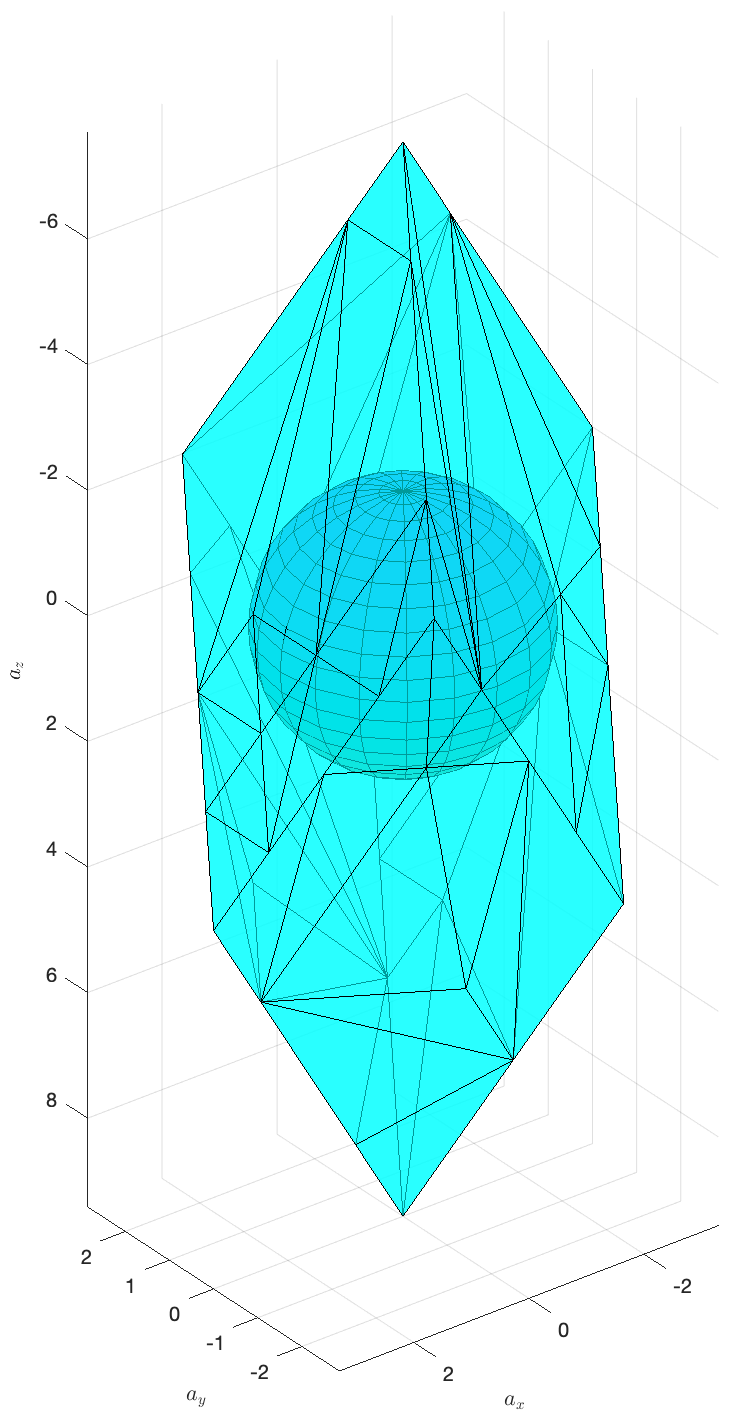}
        \caption{~}
    \end{subfigure}
    \hfill
    \begin{subfigure}[b]{0.76\linewidth}
        \includegraphics[width=\linewidth]{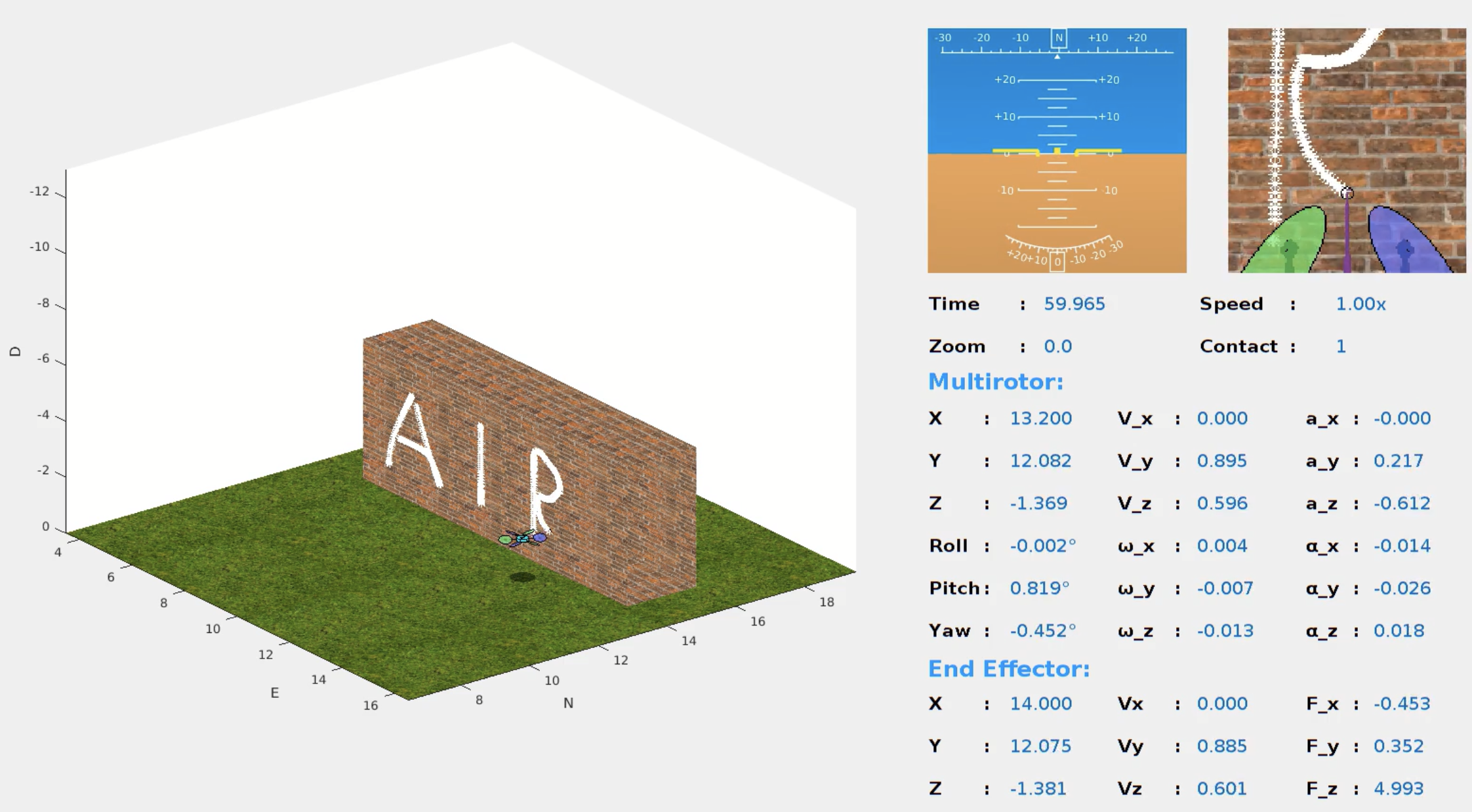}
        \caption{~}
    \end{subfigure}%
\caption{Examples of the simulator's features~\cite{Keipour:2023:scitech:simulator}. (a) A fully-actuated multirotor designed and visualized. (b) Attitude response analysis. (c) Wrench-set analysis. (d) A physical interaction application (a UAV painting on the wall).}
\label{fig:features}
\end{figure}

\begin{itemize}[leftmargin=*]
    \item Rapid aircraft modeling,
    \item Rapid controller design,
    \item Supporting physical contact of UAVs with their environment,
    \item Visualization for UAV applications,
    \item Full wrench-set analysis for any state,
    \item Logging of all internal and external variables,
    \item Simplicity of extending with new features.
\end{itemize}

Compared to Gazebo and similar simulators, our simulator significantly reduces the design and experimentation time and provides additional analysis targeted for fully-actuated UAVs and physical interaction tasks.

This workshop paper briefly discusses the major features of the ARCAD simulator. The code for the simulator can be accessed at \url{https://github.com/keipour/aircraft-simulator-matlab}. 
\section{Aircraft Design} \label{sec:model}

Our goal for developing the simulator was the ability to design different aircraft models. Therefore, the current version supports customized underactuated and fully-actuated multirotors, fixed-wing aircraft, and hybrid VTOLs with just a few lines of code. In addition to the general design of the aircraft, all parameters can be customized, such as aerodynamic properties for each rotor (e.g., the thrust and torque coefficients) and even visualization properties (e.g., the size of the rotors). Each component can also be customized; for example, each rotor can be uni- or bidirectional, fixed- or variable-pitch (with the desired axis and angle range), and a manipulation arm can be added to the robot. 

\begin{figure}[!htb]
\centering
    \begin{subfigure}[b]{0.33\linewidth}
        \includegraphics[width=\textwidth]{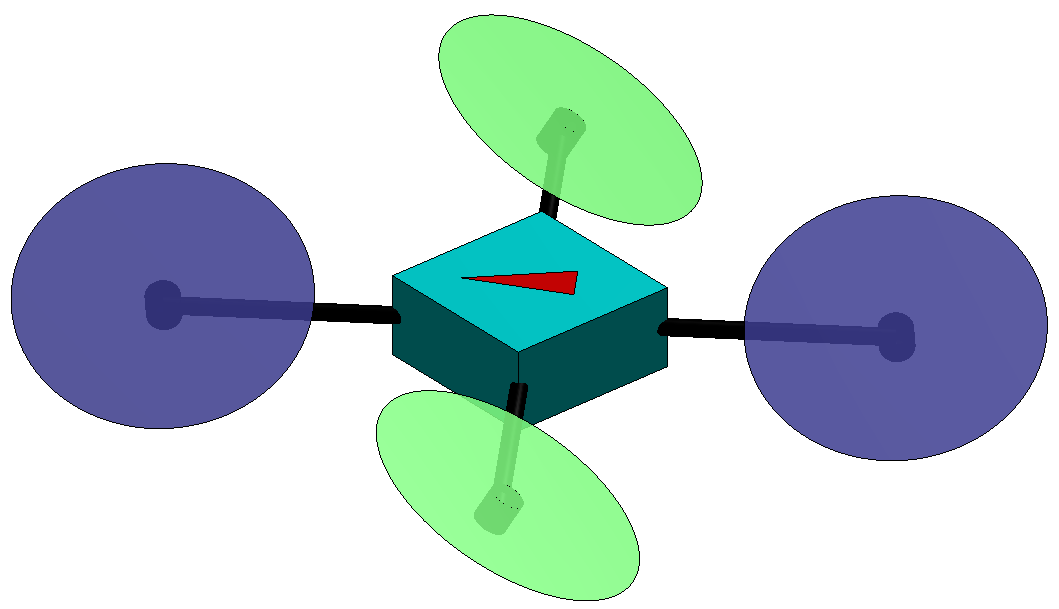}
        \caption{~}
    \end{subfigure}%
    \hfill%
    \begin{subfigure}[b]{0.33\linewidth}
        \includegraphics[width=\textwidth]{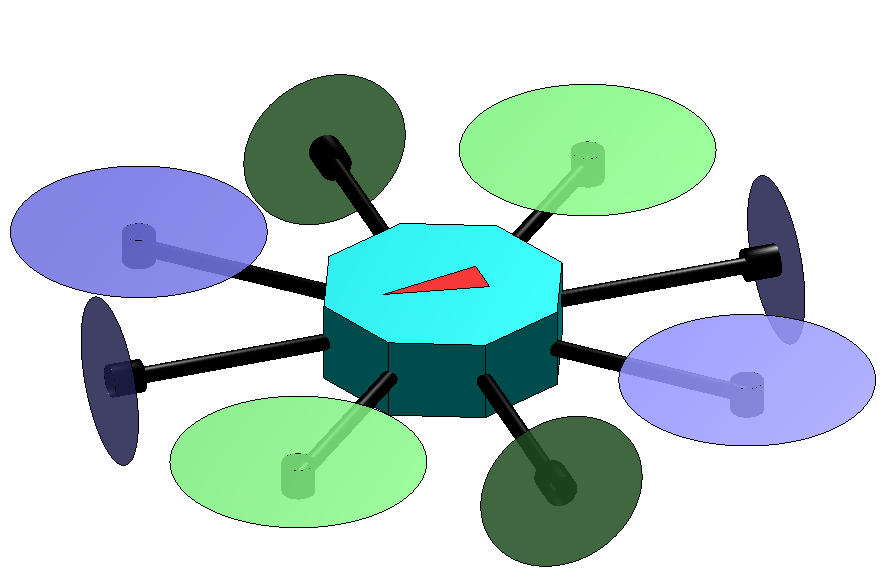}
        \caption{~}
    \end{subfigure}%
    \hfill%
    \begin{subfigure}[b]{0.33\linewidth}
        \includegraphics[width=\textwidth]{figures/tilted-hex.png}
        \caption{~}
    \end{subfigure}%
    
    \medskip
    \begin{subfigure}[b]{0.33\linewidth}
        \includegraphics[width=\textwidth]{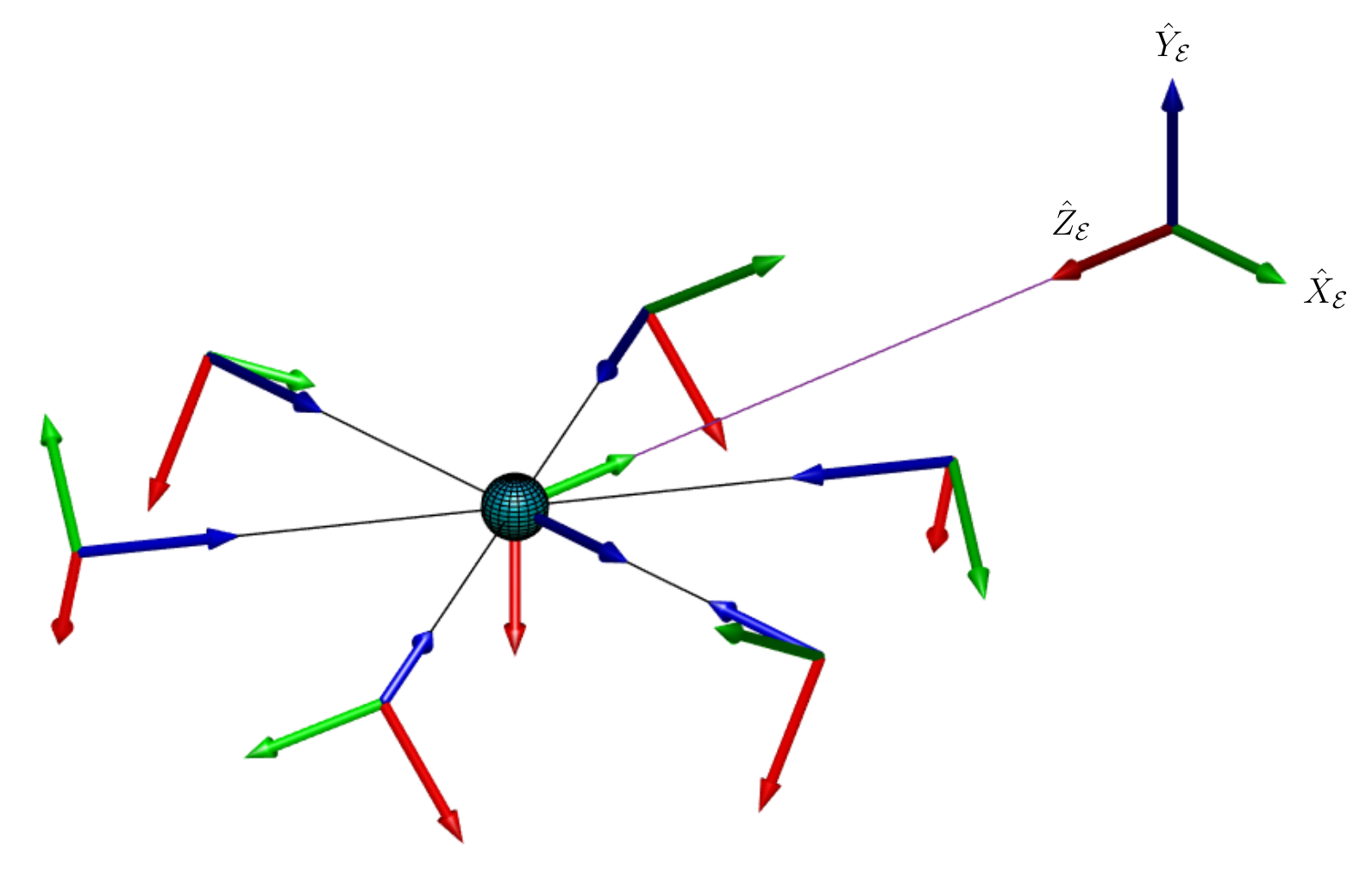}
        \caption{~}
    \end{subfigure}%
    \hfill%
    \begin{subfigure}[b]{0.33\linewidth}
        \includegraphics[width=\textwidth]{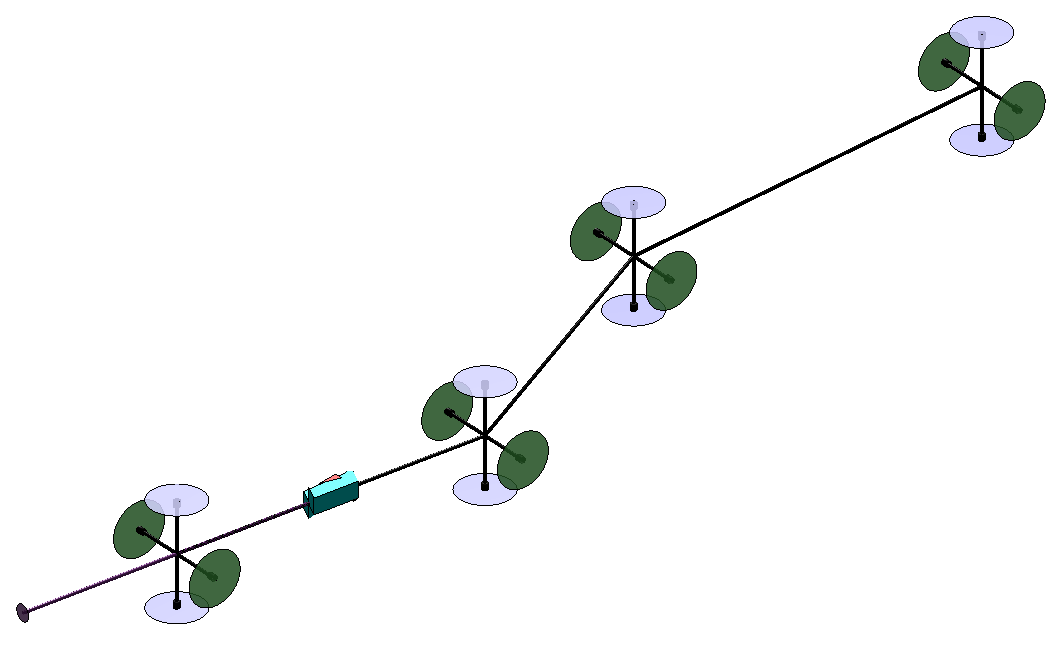}
        \caption{~}
    \end{subfigure}%
    \hfill%
    \begin{subfigure}[b]{0.33\linewidth}
        \includegraphics[width=\textwidth]{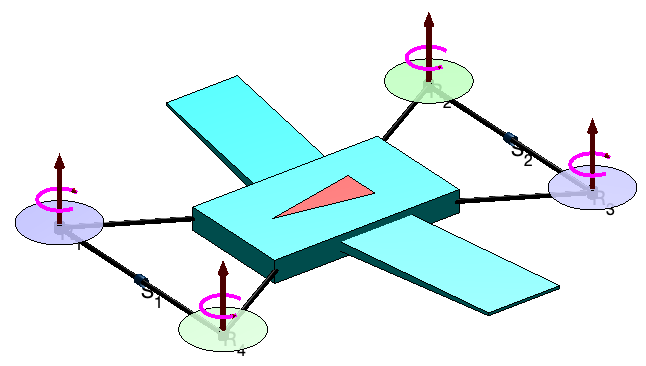}
        \caption{~}
    \end{subfigure}
\caption{Visualization of some aircraft architectures~\cite{Keipour:2023:scitech:simulator}. (a) A quadrotor with variable-pitch rotors (i.e., thrusters). (b) A fully-actuated octorotor. (c) A fully-actuated hexarotor with tilted arms and a manipulation arm, visualized with its rotor axes and rotation directions. (d) Visualization of only the axes for the hexarotors in (c). (e) A multilink multirotor aircraft. (f) A hybrid VTOL with variable pitch rotors.}
\label{fig:designs}
\end{figure}

During the design process, the 3-D visualization of the design is rendered, and all the resulting coordinate frames can be visualized to ensure they are defined as desired. Figure~\ref{fig:designs} illustrates some example designs defined in the ARCAD simulator.

\section{Controller Design} \label{sec:control}

The controller design in the simulator is modular, making it simple to swap different submodules (e.g., the attitude controller) or replace the whole controller with a new one (e.g., with an MPPI controller~\cite{Mousaei:2023:unpub:mppi}). 

Currently, there are various controllers implemented in the simulator that can be customized, ranging from simple nested PID loops for position and attitude controllers to Hybrid Position-Force controller (HPFC)~\cite{Keipour:2020:arxiv:integration, Keipour:2022:thesis} and customized controllers for fixed-wing and hybrid VTOLs~\cite{Mousaei:2022:iros:vtol, Mousaei:2022:icra-workshop:vtol}.

\begin{figure}[!htb]
\centering
\includegraphics[width=0.7\linewidth]{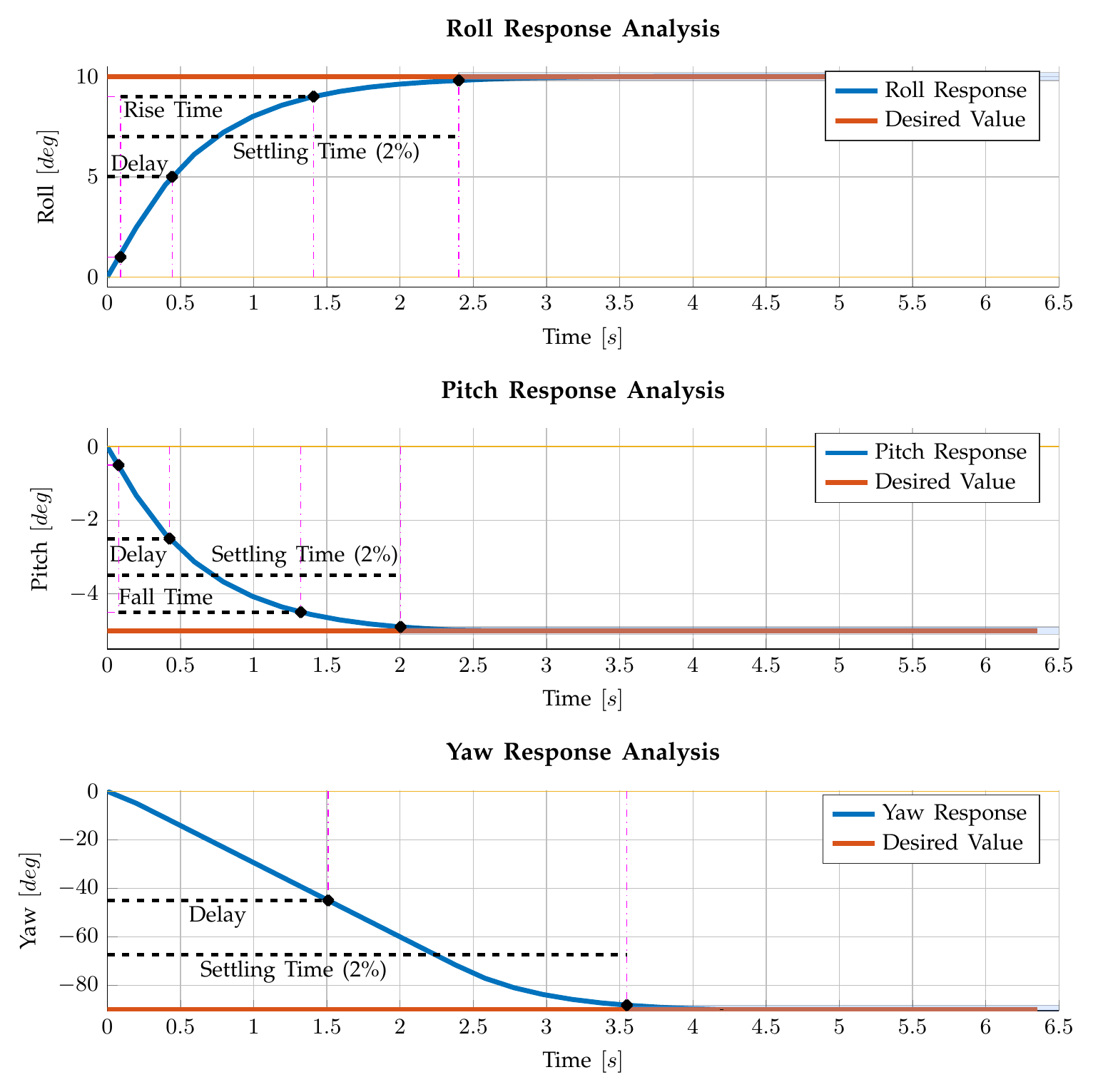}
\caption{Automatic analysis of the attitude response provided suring controller design~\cite{Keipour:2023:scitech:simulator}.}
\label{fig:attitude-response}
\end{figure}

Figure~\ref{fig:attitude-response} illustrates an example of automatic analysis provided during the controller design. 

\section{Analysis} \label{sec:analysis}

Designing fully-actuated multirotors and new aircraft types requires analysis beyond the limited requirements for underactuated designs. The ARCAD simulator provides real-time analysis for the wrench set~\cite{Keipour:2023:unpub:wrench} of the design and supports visualization of all desired signals to facilitate model and controller design as well as developing more advanced UAV applications. 

\begin{figure}[!htb]
\centering
    \begin{subfigure}[b]{0.38\linewidth}
        \includegraphics[width=\textwidth]{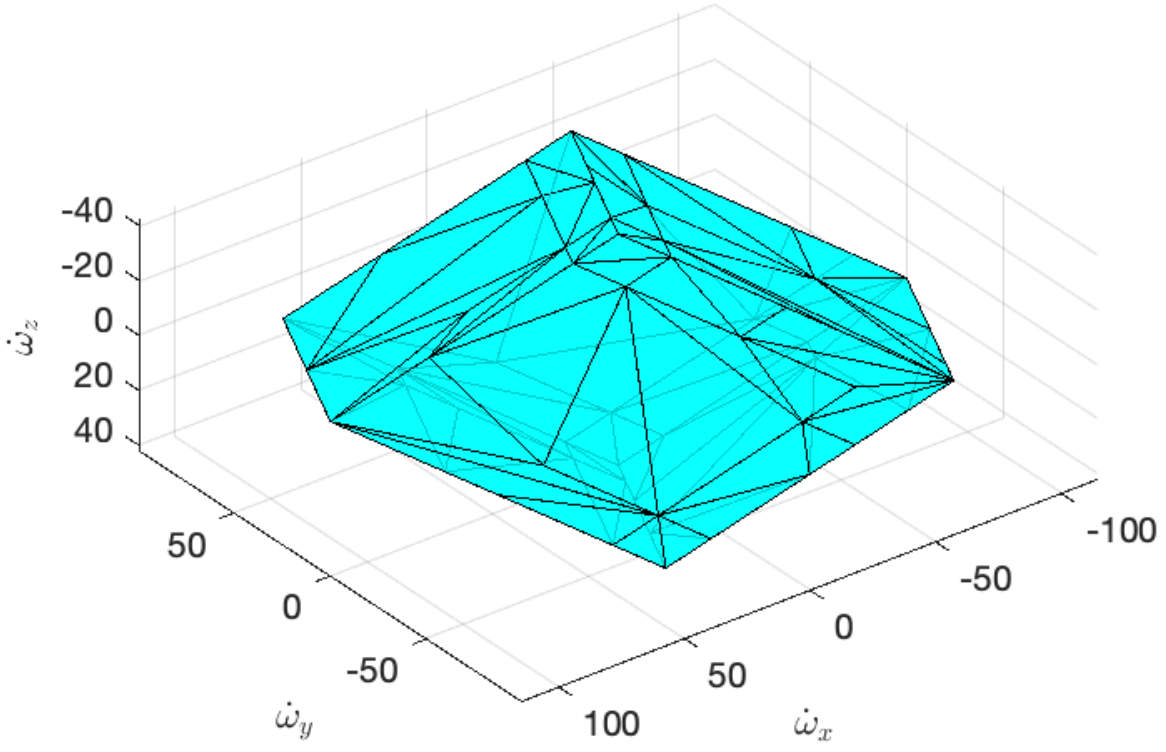}
        \caption{~}
    \end{subfigure}%
    \hfill%
    \begin{subfigure}[b]{0.16\linewidth}
        \includegraphics[width=\textwidth]{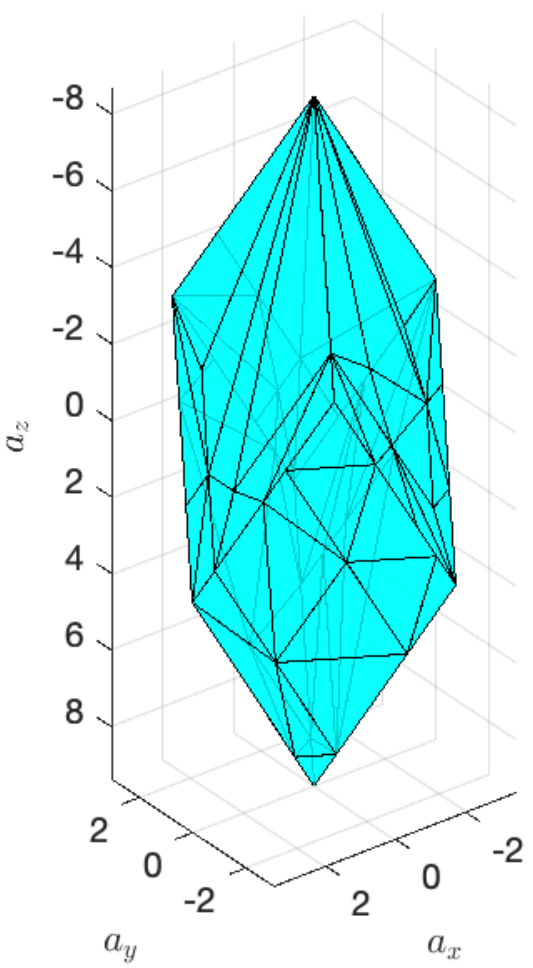}
        \caption{~}
    \end{subfigure}%
    \hfill%
    \begin{subfigure}[b]{0.38\linewidth}
        \includegraphics[width=\textwidth]{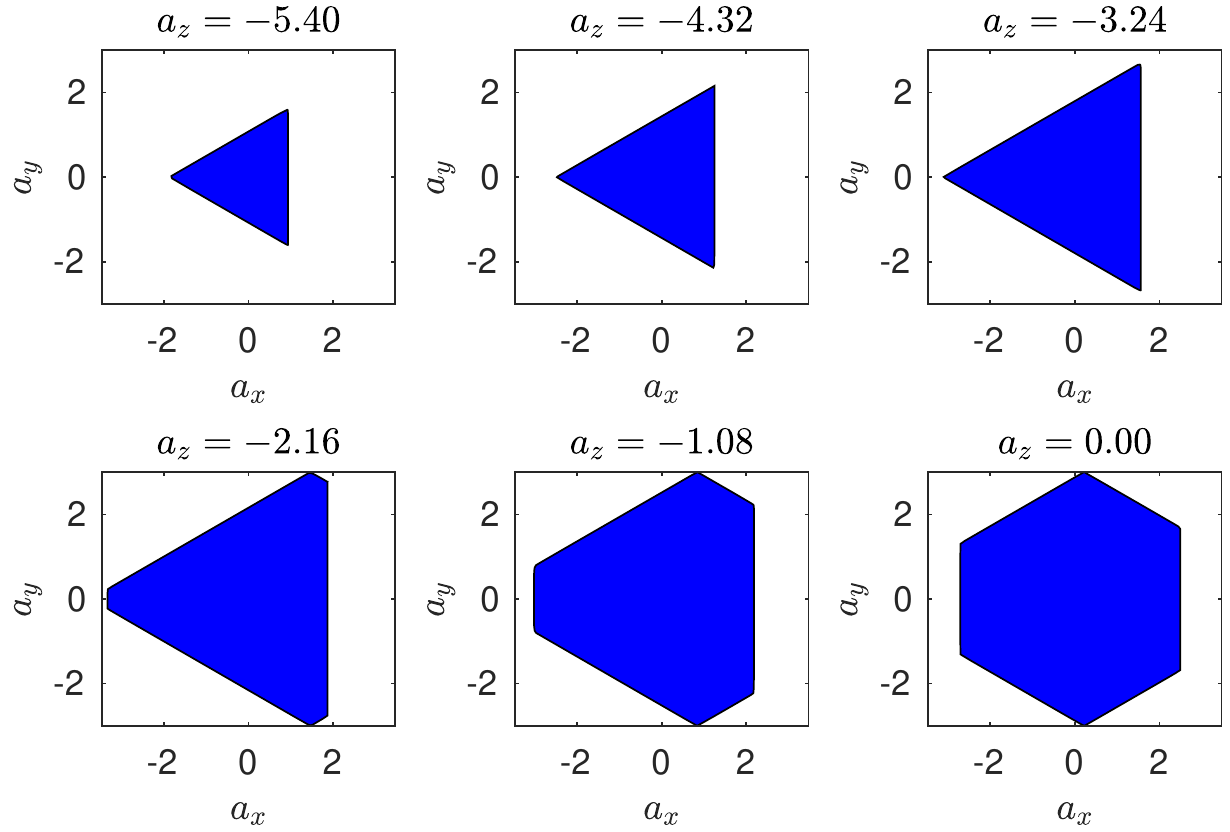}
        \caption{~}
    \end{subfigure}%
    
    \medskip
    
    \begin{subfigure}[b]{0.38\linewidth}
        \includegraphics[width=\textwidth]{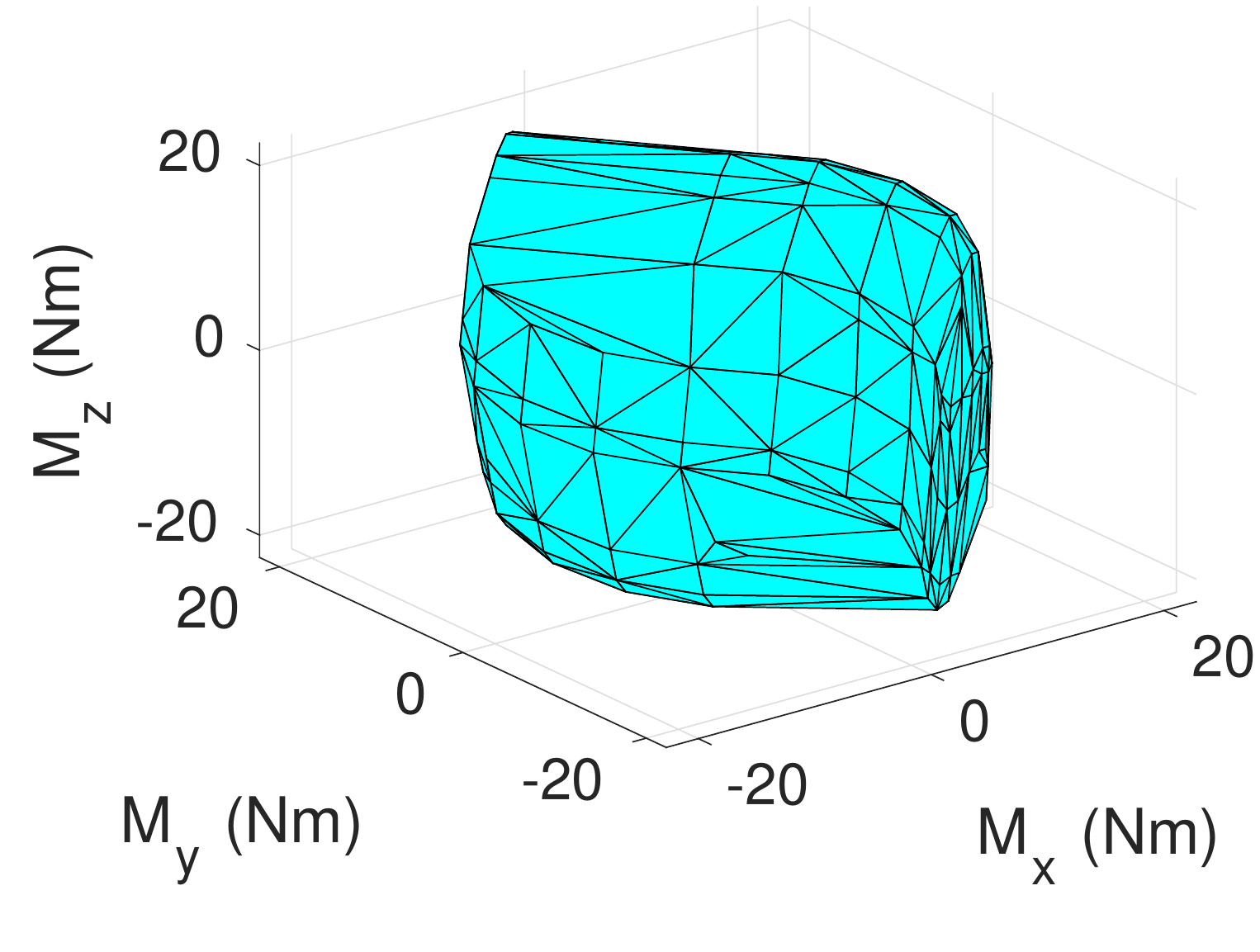}
        \caption{~}
    \end{subfigure}%
    \hfill
    \begin{subfigure}[b]{0.16\linewidth}
        \includegraphics[width=\textwidth]{figures/omni-directional-acceleration-sphere.png}
        \caption{~}
    \end{subfigure}%
    \hfill%
    \begin{subfigure}[b]{0.38\linewidth}
        \includegraphics[width=\textwidth]{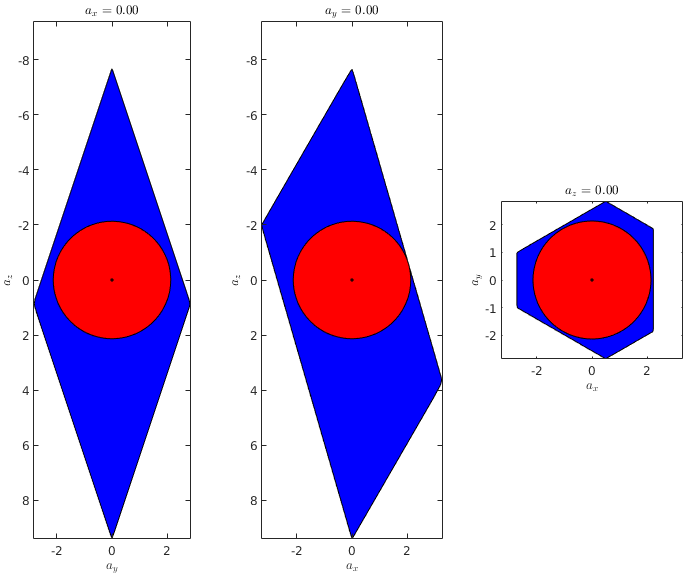}
        \caption{~}
    \end{subfigure}%

    \caption{Wrench set and omni-directional acceleration analysis for aircraft~\cite{Keipour:2023:scitech:simulator}. (a) Moment set for a fully-actuated multirotor. (b, c) Force set and cross-sections for the same multirotor. (d) Moment set for the hybrid VTOL. (e, f) Omni-directional acceleration sphere and cross-sections (along all three axes) for the hexarotor with tilted arms.}
\label{fig:analysis}
\end{figure}

\begin{figure}[!htb]
\centering
\includegraphics[width=\linewidth]{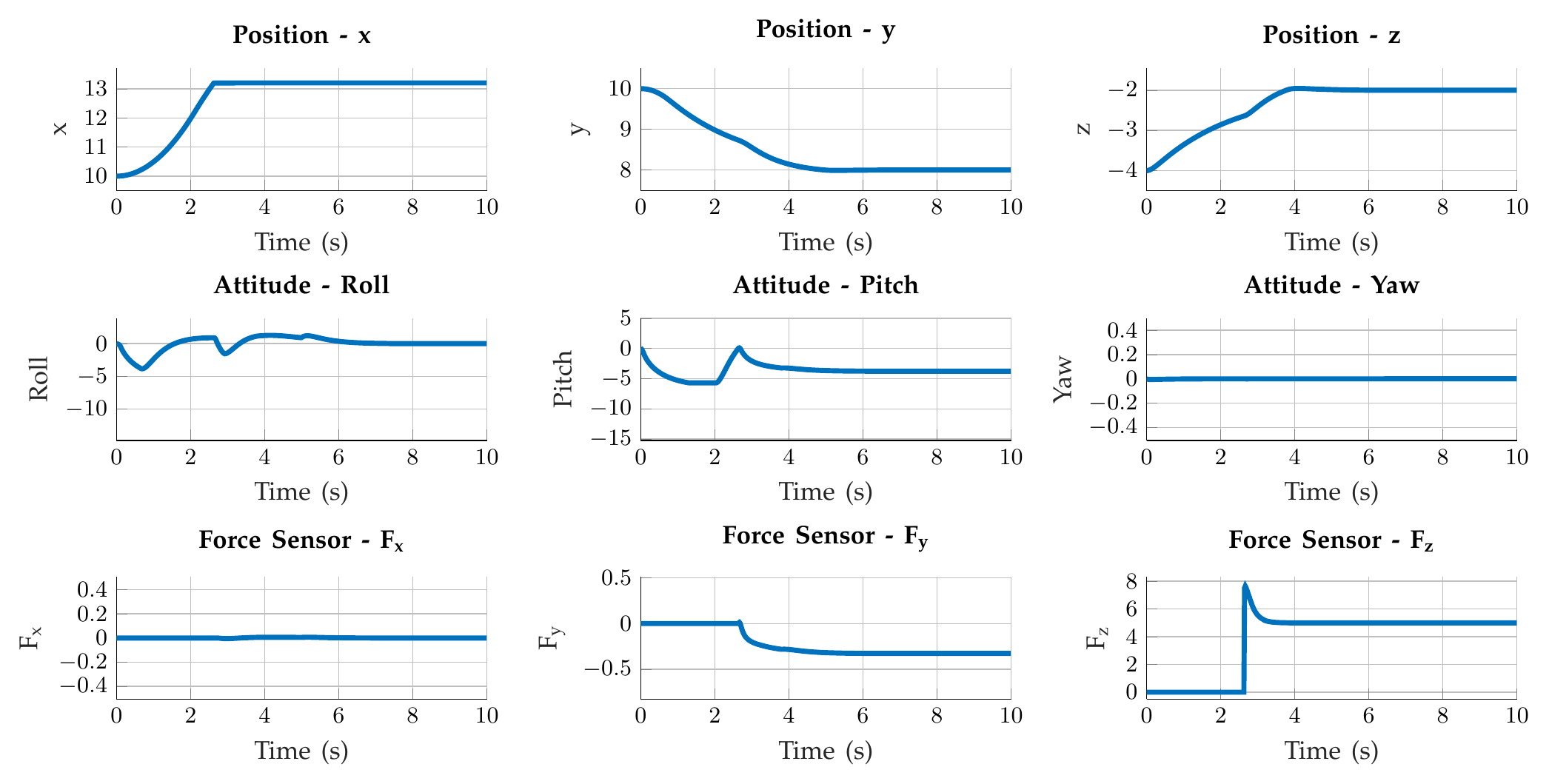}
\caption{A sample plot generated for a multirotor applying a 5~N force to a wall~\cite{Keipour:2023:scitech:simulator}.}
\label{fig:sample-plots}
\end{figure}

Figure~\ref{fig:analysis} shows an example wrench set analysis for the aircraft, and Figure~\ref{fig:sample-plots} illustrates the sample plots for a physical interaction application.

\section{Applications} \label{sec:applications}

The goal of developing a new simulator from scratch was to enable our research on controller design and aircraft model analysis. However, over time additional features were implemented that have enabled the development of applications for the designed aircraft. Features such as 3-D and POV camera views, devices such as force/torque sensors, 3-D obstacle definitions with collision models, GUIs, higher-level controllers, and high-quality video recordings of the simulation are just a few examples.

\begin{figure}[!htb]
\centering
\includegraphics[width=\linewidth]{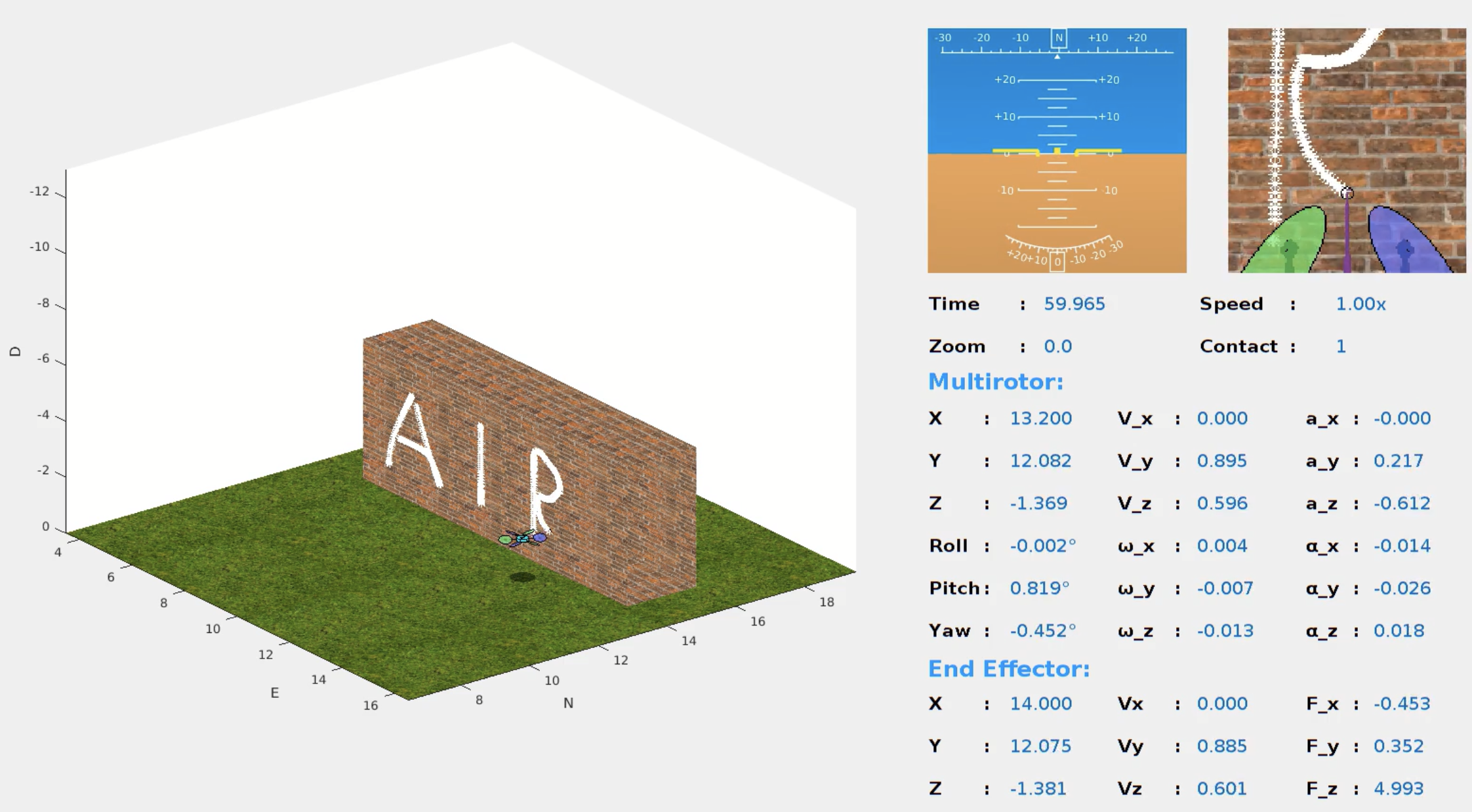}
\caption{A multirotor writing AIR on the wall while controlling the applied force to 5~N~\cite{Keipour:2023:scitech:simulator}.}
\label{fig:applications}
\end{figure}

\begin{figure}[!htb]
\centering
\includegraphics[width=0.8\linewidth]{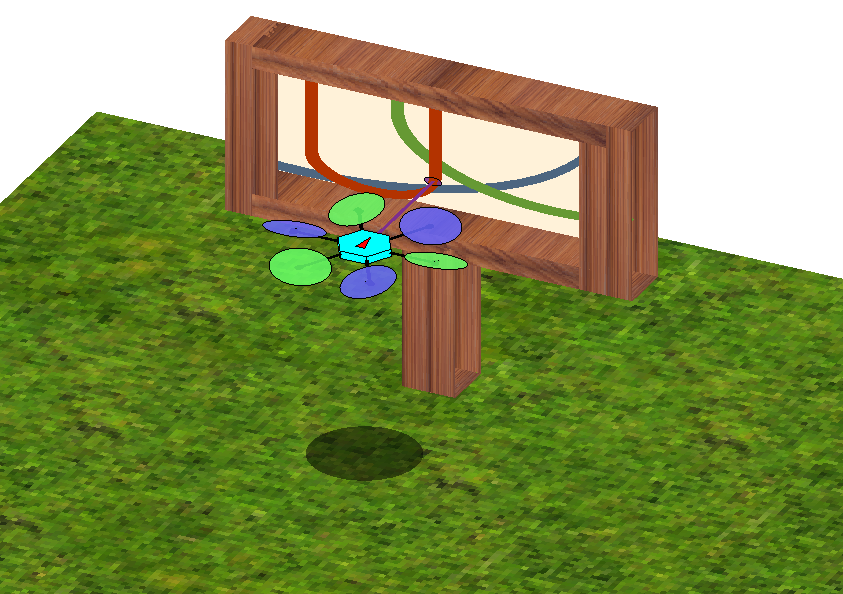}
\caption{An application for wire manipulation using UAVs~\cite{Keipour:2022:icra-workshop:doo}.}
\label{fig:wire-manip}
\end{figure}

Figure~\ref{fig:applications} shows a snapshot of the application where the UAV is writing on the wall with a controlled force, and Figure~\ref{fig:wire-manip} shows a wire manipulation application. 

\section{Conclusion} \label{sec:conclusion}

This workshop paper presented the ARCAD simulator that has allowed our team rapidly prototype and develop new fully-actuated aircraft, controllers, and applications, including physical interaction. 

Currently, the simulator is in MATLAB, which due to historical reasons, was the right choice for our development at the time. However, we realize that compared to other prototyping languages, such as Python, MATLAB can limit public access due to licensing and has lower performance. Additionally, due to the recent popularity spike in Machine Learning, Python has become a better choice for the public interest. Therefore, we will slowly transition the simulator to Python over time and improve the documentation. Moreover, in future iterations, we plan to integrate the simulator with ROS and MAVLINK. 

Several real-world controllers and applications in our lab have been developed in this simulator and were deployed on real robots only after improving and verifying the methods. We hope this simulator can accelerate the research on fully-actuated and other new types of aircraft and support the development of new applications. The simulator can be accessed from \url{https://github.com/keipour/aircraft-simulator-matlab}.

\addtolength{\textheight}{-14.0cm}   





\bibliographystyle{IEEEtran}
\bibliography{paper-citations.bib}

\begin{thebibliography}{10}
\providecommand{\url}[1]{#1}
\csname url@rmstyle\endcsname
\providecommand{\newblock}{\relax}
\providecommand{\bibinfo}[2]{#2}
\providecommand\BIBentrySTDinterwordspacing{\spaceskip=0pt\relax}
\providecommand\BIBentryALTinterwordstretchfactor{4}
\providecommand\BIBentryALTinterwordspacing{\spaceskip=\fontdimen2\font plus
\BIBentryALTinterwordstretchfactor\fontdimen3\font minus
  \fontdimen4\font\relax}
\providecommand\BIBforeignlanguage[2]{{%
\expandafter\ifx\csname l@#1\endcsname\relax
\typeout{** WARNING: IEEEtran.bst: No hyphenation pattern has been}%
\typeout{** loaded for the language `#1'. Using the pattern for}%
\typeout{** the default language instead.}%
\else
\language=\csname l@#1\endcsname
\fi
#2}}

\bibitem{Balaram2021}
\BIBentryALTinterwordspacing
J.~Balaram, M.~Aung, and M.~P. Golombek, ``The ingenuity helicopter on the
  perseverance rover,'' \emph{Space Science Reviews}, vol. 217, no.~4, p.~56,
  May 2021. [Online]. Available:
  \url{https://doi.org/10.1007/s11214-021-00815-w}
\BIBentrySTDinterwordspacing

\bibitem{en15010217}
\BIBentryALTinterwordspacing
P.~Velusamy, S.~Rajendran, R.~K. Mahendran, S.~Naseer, M.~Shafiq, and J.-G.
  Choi, ``Unmanned aerial vehicles (uav) in precision agriculture: Applications
  and challenges,'' \emph{Energies}, vol.~15, no.~1, 2022. [Online]. Available:
  \url{https://www.mdpi.com/1996-1073/15/1/217}
\BIBentrySTDinterwordspacing

\bibitem{Keipour:2022:sensors:mbzirc}
\BIBentryALTinterwordspacing
A.~Keipour, G.~A.~S. Pereira, R.~Bonatti, R.~Garg, P.~Rastogi, G.~Dubey, and
  S.~A. Scherer, ``Visual servoing approach for autonomous {UAV} landing on a
  moving vehicle,'' \emph{Sensors, Special Issue: Advanced Sensors Technologies
  Applied in Mobile Robot}, vol.~22, no.~17, pp. 1--18, 8 2022. [Online].
  Available: \url{https://www.mdpi.com/1424-8220/22/17/6549}
\BIBentrySTDinterwordspacing

\bibitem{RAKHA2018252}
\BIBentryALTinterwordspacing
T.~Rakha and A.~Gorodetsky, ``Review of unmanned aerial system (uas)
  applications in the built environment: Towards automated building inspection
  procedures using drones,'' \emph{Automation in Construction}, vol.~93, pp.
  252--264, 2018. [Online]. Available:
  \url{https://www.sciencedirect.com/science/article/pii/S0926580518300165}
\BIBentrySTDinterwordspacing

\bibitem{Mousaei:2022:iros:vtol}
\BIBentryALTinterwordspacing
M.~Mousaei, J.~Geng, A.~Keipour, D.~Bai, and S.~Scherer, ``Design, modeling and
  control for a tilt-rotor {VTOL} {UAV} in the presence of actuator failure,''
  in \emph{2022 IEEE/RSJ International Conference on Intelligent Robots and
  Systems (IROS)}, Kyoto, Japan, 10 2022, pp. 4310--4317. [Online]. Available:
  \url{https://ieeexplore.ieee.org/document/9981806}
\BIBentrySTDinterwordspacing

\bibitem{Park2018a}
\BIBentryALTinterwordspacing
S.~Park, J.~Lee, J.~Ahn, M.~Kim, J.~Her, G.-H. Yang, and D.~Lee, ``Odar: Aerial
  manipulation platform enabling omnidirectional wrench generation,''
  \emph{IEEE/ASME Transactions on Mechatronics}, vol.~23, no.~4, pp.
  1907--1918, aug 2018. [Online]. Available:
  \url{https://ieeexplore.ieee.org/document/8401328/}
\BIBentrySTDinterwordspacing

\bibitem{Rashad2019a}
\BIBentryALTinterwordspacing
R.~Rashad, F.~Califano, and S.~Stramigioli, ``Port-hamiltonian passivity-based
  control on {SE(3)} of a fully actuated {UAV} for aerial physical interaction
  near-hovering,'' \emph{IEEE Robotics and Automation Letters}, vol.~4, no.~4,
  pp. 4378--4385, oct 2019. [Online]. Available:
  \url{https://ieeexplore.ieee.org/document/8786163/}
\BIBentrySTDinterwordspacing

\bibitem{Trujillo2019}
\BIBentryALTinterwordspacing
M.~{\'{A}}. {Trujillo Soto}, J.~{Mart{\'{i}}nez-de Dios}, C.~Mart{\'{i}}n,
  A.~{Viguria Jimenez}, and A.~Ollero, ``\BIBforeignlanguage{eng}{Novel aerial
  manipulator for accurate and robust industrial {NDT} contact inspection: A
  new tool for the oil and gas inspection industry},''
  \emph{\BIBforeignlanguage{eng}{Sensors}}, vol.~19, no.~6, p. 1305, mar 2019.
  [Online]. Available: \url{https://pubmed.ncbi.nlm.nih.gov/30875905}
\BIBentrySTDinterwordspacing

\bibitem{Ollero2018}
\BIBentryALTinterwordspacing
A.~Ollero, J.~Cortes, A.~Santamaria-Navarro, M.~{\'{A}}. {Trujillo Soto},
  R.~Balachandran, J.~Andrade-Cetto, A.~Rodriguez, G.~Heredia, A.~Franchi,
  G.~Antonelli, K.~Kondak, A.~Sanfeliu, A.~{Viguria Jimenez}, J.~R.
  {Martinez-de Dios}, and F.~Pierri, ``The aeroarms project: Aerial robots with
  advanced manipulation capabilities for inspection and maintenance,''
  \emph{IEEE Robotics {\&} Automation Magazine}, vol.~25, no.~4, pp. 12--23,
  dec 2018. [Online]. Available:
  \url{https://ieeexplore.ieee.org/document/8435987/}
\BIBentrySTDinterwordspacing

\bibitem{Keipour:2020:arxiv:integration}
\BIBentryALTinterwordspacing
A.~Keipour, M.~Mousaei, A.~T. Ashley, and S.~Scherer, ``Integration of
  fully-actuated multirotors into real-world applications,''
  \emph{arXiv:2011.06666}, pp. 1--5, 2020. [Online]. Available:
  \url{https://arxiv.org/abs/2011.06666}
\BIBentrySTDinterwordspacing

\bibitem{2012simpar_meyer}
J.~Meyer, A.~Sendobry, S.~Kohlbrecher, U.~Klingauf, and O.~von Stryk,
  ``Comprehensive simulation of quadrotor {UAV}s using {ROS} and {Gazebo},'' in
  \emph{Simulation, Modeling, and Programming for Autonomous Robots}, I.~Noda,
  N.~Ando, D.~Brugali, and J.~J. Kuffner, Eds.\hskip 1em plus 0.5em minus
  0.4em\relax Berlin, Heidelberg: Springer Berlin Heidelberg, 2012, pp.
  400--411.

\bibitem{airsim2017fsr}
S.~Shah, D.~Dey, C.~Lovett, and A.~Kapoor, ``Airsim: High-fidelity visual and
  physical simulation for autonomous vehicles,'' in \emph{Field and Service
  Robotics}, M.~Hutter and R.~Siegwart, Eds.\hskip 1em plus 0.5em minus
  0.4em\relax Cham: Springer International Publishing, 2018, pp. 621--635.

\bibitem{Furrer2016}
\BIBentryALTinterwordspacing
F.~Furrer, M.~Burri, M.~Achtelik, and R.~Siegwart, \emph{Robot Operating System
  (ROS): The Complete Reference (Volume 1)}.\hskip 1em plus 0.5em minus
  0.4em\relax Cham: Springer International Publishing, 2016, ch. RotorS---A
  Modular Gazebo MAV Simulator Framework, pp. 595--625. [Online]. Available:
  \url{http://dx.doi.org/10.1007/978-3-319-26054-9_23}
\BIBentrySTDinterwordspacing

\bibitem{perry2004flightgear}
A.~R. Perry, ``The {FlightGear} flight simulator,'' in \emph{Proceedings of the
  USENIX Annual Technical Conference}, vol. 686, 2004, pp. 1--12.

\bibitem{x_2022}
\BIBentryALTinterwordspacing
``{X-plane~12},'' Nov 2022. [Online]. Available: \url{https://www.x-plane.com/}
\BIBentrySTDinterwordspacing

\bibitem{Keipour:2023:scitech:simulator}
\BIBentryALTinterwordspacing
A.~Keipour, M.~Mousaei, D.~Bai, J.~Geng, and S.~Scherer, ``{UAS} simulator for
  modeling, analysis and control in free flight and physical interaction,'' in
  \emph{AIAA SciTech 2023 Forum}.\hskip 1em plus 0.5em minus 0.4em\relax
  National Harbor, MD, USA: American Institute of Aeronautics and Astronautics,
  1 2023. [Online]. Available:
  \url{https://arc.aiaa.org/doi/abs/10.2514/6.2023-1279}
\BIBentrySTDinterwordspacing

\bibitem{Keipour:2022:thesis}
\BIBentryALTinterwordspacing
A.~Keipour, ``Physical interaction and manipulation of the environment using
  aerial robots,'' Ph.D. dissertation, Carnegie Mellon University, Pittsburgh,
  PA, USA, 5 2022. [Online]. Available: \url{https://arxiv.org/abs/2207.02856}
\BIBentrySTDinterwordspacing

\bibitem{Mousaei:2023:unpub:mppi}
M.~Mousaei, J.~Geng, A.~Keipour, and S.~Scherer, ``A unified control framework
  for aerial manipulation with large maneuverability,'' 5 2023, in press.

\bibitem{Mousaei:2022:icra-workshop:vtol}
\BIBentryALTinterwordspacing
M.~Mousaei, A.~Keipour, J.~Geng, and S.~Scherer, ``{VTOL} failure detection and
  recovery by utilizing redundancy,'' in \emph{Workshop on Intelligent Aerial
  Robotics: From Autonomous Micro Aerial Vehicles to Sustainable Urban Air
  Mobility and Operations, 2022 IEEE International Conference on Robotics and
  Automation (ICRA)}, Philadelphia, PA, USA, 5 2022, pp. 1--5. [Online].
  Available: \url{https://arxiv.org/abs/2206.00588}
\BIBentrySTDinterwordspacing

\bibitem{Keipour:2023:unpub:wrench}
A.~Keipour, M.~Mousaei, J.~Geng, and S.~Scherer, ``Real-time wrench-set
  analysis and applications for aerial robots,'' 8 2023, in press.

\bibitem{Keipour:2022:icra-workshop:doo}
\BIBentryALTinterwordspacing
A.~Keipour, M.~Mousaei, M.~Bandari, S.~Schaal, and S.~Scherer, ``Detection and
  physical interaction with deformable linear objects,'' in \emph{2nd Workshop
  on Representing and Manipulating Deformable Objects, 2022 IEEE International
  Conference on Robotics and Automation (ICRA)}, Philadelphia, PA, USA, 5 2022,
  pp. 1--4. [Online]. Available: \url{https://arxiv.org/abs/2205.08041}
\BIBentrySTDinterwordspacing

\end{thebibliography}

\end{document}